\documentclass[10pt,twocolumn,letterpaper]{article}

\usepackage{iccv}
\usepackage{times}
\usepackage{epsfig}
\usepackage{graphicx}
\usepackage{amsmath}
\usepackage{amssymb}
\usepackage{cuted}
\usepackage{capt-of}
\usepackage{animate}
\usepackage{subfigure}
\usepackage{caption}

\def \useAnimate {}

\usepackage[pagebackref=true,breaklinks=true,letterpaper=true,colorlinks,bookmarks=false]{hyperref}

\iccvfinalcopy 

\def\iccvPaperID{30} 

\ificcvfinal\fi
\begin{document}

\title{Controllable Artistic Text Style Transfer via Shape-Matching GAN\vspace{-2mm}}

\author{Shuai Yang$^{1,2}$,~Zhangyang Wang$^2$,~Zhaowen Wang$^3$,~Ning Xu$^3$,~Jiaying Liu\thanks{Corresponding author\newline The work was done when Shuai Yang was a visiting student at TAMU.}~$^1$~and Zongming Guo$^1$\\
$^1$ Institute of Computer Science and Technology, Peking University\\
$^2$ Texas A\&M University~~~~$^3$ Adobe Research\vspace{-6mm}
}

\maketitle
\ificcvfinal\thispagestyle{empty}\fi
\captionsetup[subfigure]{labelsep=space}

\begin{strip}\centering
\noindent\begin{minipage}{\textwidth}
\begin{minipage}{0.134\textwidth}
  \centering
  \includegraphics[width=.98\textwidth]{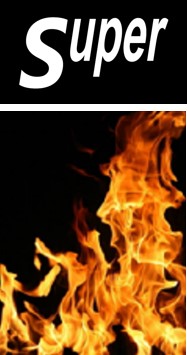}\vspace{-2mm}
  \captionof{subfigure}{source image}
\end{minipage}%
\begin{minipage}{0.333\textwidth}
  \centering
  \includegraphics[width=0.98\textwidth]{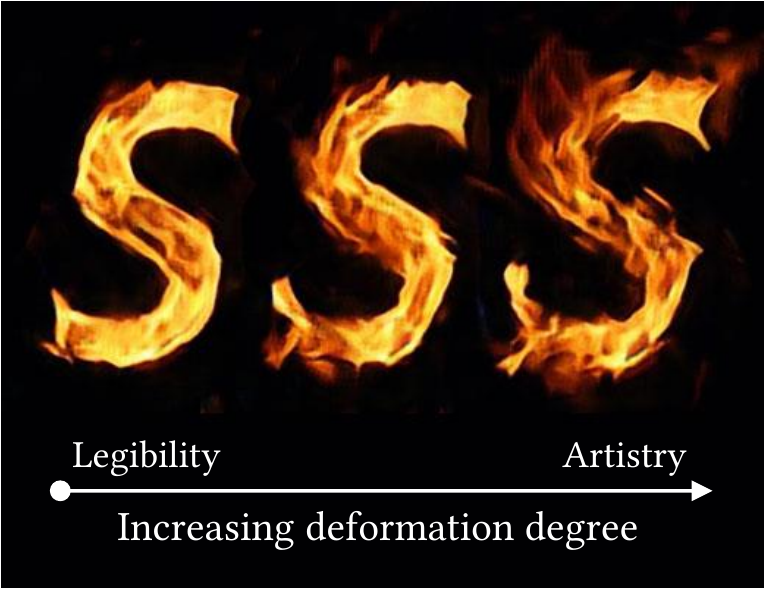}\vspace{-2mm}
  \captionof{subfigure}{adjustable stylistic degree of glyph}
\end{minipage}%
\begin{minipage}{0.341\textwidth}
  \centering
  \ifx \useAnimate \undefined
    \includegraphics[width=0.98\textwidth]{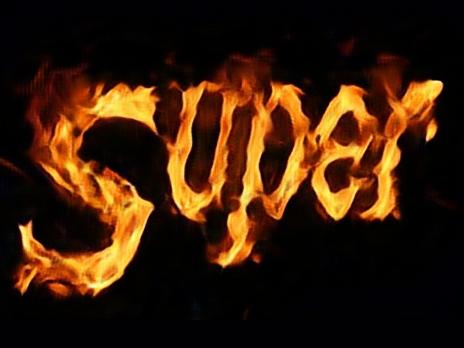}\vspace{-2mm}
  \else
    \animategraphics[width=0.98\textwidth,autoplay,loop]{12}{figures/teaser/new/texture_}{0}{49}\vspace{-2mm}
  \fi
  \captionof{subfigure}{stylized text}
\end{minipage}%
\begin{minipage}{0.181\textwidth}
  \centering
  \ifx \useAnimate \undefined
    \includegraphics[width=0.98\textwidth]{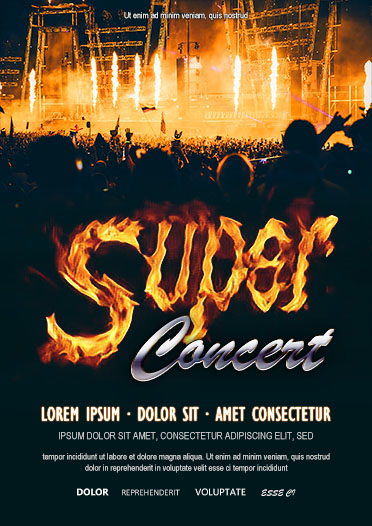}\vspace{-2mm}
  \else
    \animategraphics[width=0.98\textwidth,autoplay,loop]{4}{figures/teaser/poster/texture_}{0}{8}\vspace{-2mm}
  \fi
  \captionof{subfigure}{application}
\end{minipage}\par\vspace{1mm}
\begin{minipage}{0.498\textwidth}
  \centering
  \ifx \useAnimate \undefined
    \includegraphics[width=.98\textwidth]{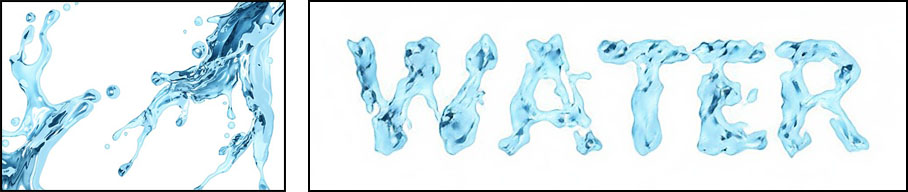}\vspace{-2mm}
  \else
    \animategraphics[width=0.98\textwidth,autoplay,loop]{8}{figures/teaser/water/texture_}{0}{16}\vspace{-2mm}
  \fi
  \captionof{subfigure}{liquid artistic text rendering}
\end{minipage}%
\begin{minipage}{0.498\textwidth}
  \centering
  \ifx \useAnimate \undefined
    \includegraphics[width=.98\textwidth]{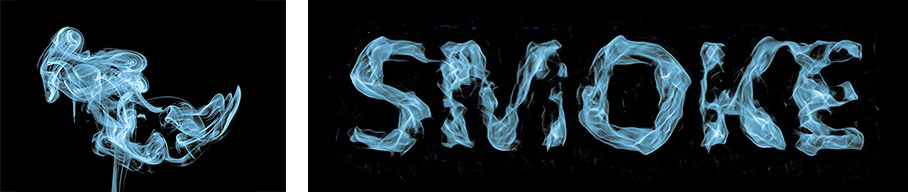}\vspace{-2mm}
  \else
    \animategraphics[width=0.98\textwidth,autoplay,loop]{8}{figures/teaser/smoke/texture_}{0}{16}\vspace{-2mm}
  \fi
  \captionof{subfigure}{smoke artistic text rendering}
\end{minipage}\vspace{-2mm}
\captionof{figure}{We propose a novel style transfer framework for rendering artistic text from a source style image in a scale-controllable manner.
Our framework allows users to (b) adjust the stylistic degree of the glyph (\textit{i.e.} deformation degree) in a continuous and real-time way, and therefore to (c) select the artistic text that is most ideal for both legibility and style consistency.
The generated diverse artistic text will facilitate users to design (d) exquisite posters and (e)(f) dynamic typography.
\textit{Embedded animation best viewed in Acrobat Reader.}}
\label{fig:teaser}
\end{minipage}
\end{strip}

\begin{abstract}
   Artistic text style transfer is the task of migrating the style from a source image to the target text to create artistic typography. Recent style transfer methods have considered texture control to enhance usability. However, controlling the stylistic degree in terms of shape deformation remains an important open challenge. In this paper, we present the first text style transfer network that allows for real-time control of the crucial stylistic degree of the glyph through an adjustable parameter. Our key contribution is a novel bidirectional shape matching framework to establish an effective glyph-style mapping at various deformation levels without paired ground truth. Based on this idea, we propose a scale-controllable module to empower a single network to continuously characterize the multi-scale shape features of the style image and transfer these features to the target text. The proposed method demonstrates its superiority over previous state-of-the-arts in generating diverse, controllable and high-quality stylized text.
\end{abstract}

\section{Introduction}
\vspace{-0.5em}
Artistic text style transfer aims to render text in the style specified by a reference image, which is widely desired in many visual creation tasks such as poster and advertisement design.
Depending on the reference image, text can be stylized either by making analogy of existing well-designed text effects~\cite{Yang2017Awesome}, or by imitating the visual features from more general free-form style images~\cite{Yang2018Context}: the latter provides more flexibility and creativity.

For general style images as reference, since text is significantly different from and more structured than natural images, more attention should be paid to its stroke shape in the stylization of text. For example, one needs to manipulate the stylistic degree or shape deformations of a glyph to resemble the style subject \textrm{flames} in Fig.~\ref{fig:teaser}(b).
Meanwhile, the glyph legibility needs to be maintained so that the stylized text is still recognizable.
Such a delicate balance is subjective and hard to attain automatically. Therefore, a practical tool allowing users to control the stylistic degree of the glyph is of great value. Further, as users are prone to trying various settings before obtaining desired effects, real-time response to online adjustment is important.



In the literature, some efforts have been devoted to addressing  \textit{fast scale-controllable style transfer}. They trained fast feed-forward networks, with the main focus on the scale of textures like the texture strength ~\cite{Babaeizadeh2018AdjustableRS}, or the size of texture patterns~\cite{jing2018stroke}. Up to our best knowledge, there has been no work discussing \textbf{the real-time control of glyph deformations}, which is rather crucial for text style transfer.




In view of the above, we are motivated to investigate a new problem of fast controllable artistic text style transfer from a single style image. We aim at the real-time adjustment for the stylistic degree of the glyph in terms of shape deformations. It can allow users to navigate around different forms of the rendered text and select the most desired one, as illustrated in Fig.~\ref{fig:teaser}(b)(c). The challenges of fast controllable artistic text style transfer lie in two aspects. \underline{On one hand}, in contrast to well-defined scales such as the texture strength that can be straightforwardly modelled by hyper-parameters, the glyph deformation degree is subjective, neither clearly defined nor easy to parameterize. \underline{On the other hand},
there does not exist a large-scale \textit{paired} training set with both source text images and the corresponding results stylized (deformed) in different degrees. Usually, only one reference image is available for a certain style. It is thus also not straightforward to train data-driven models to learn multi-scale glyph stylization.




In this work, we propose a novel Shape-Matching GAN to address these challenges.
Our key idea is a bidirectional shape matching strategy to establish the shape mapping between source styles and target glyphs through both backward and forward transfers.
We first show that the glyph deformation can be modelled as a coarse-to-fine shape mapping of the style image, where the deformation degree is controlled by the coarse level.
Based on this idea, we develop a sketch module that simplifies the style image to various coarse levels by
backward transferring the shape features from the text to the style image.
Resulting coarse-fine image pairs provide a robust multi-scale shape mapping for data-driven learning.
With this obtained data, we build a scale-controllable module, \textit{Controllable ResBlock}, that empowers the network to learn to characterize and infer the style features on a continuous scale from the mapping.
Eventually, we can forward transfer the features of any specified scale to target glyphs to achieve scale-controllable style transfer.
In summary, our contributions are threefold:
\begin{itemize} \vspace{-2mm}
  \item We investigate the new problem of fast controllable artistic text style transfer, in terms of glyph deformations, and propose a novel bidirectional shape matching framework to solve it.\vspace{-2mm}
  \item We develop a sketch module to match the shape from the style to the glyph, which transforms a single style image to paired training data at various scales and thus enables learning robust glyph-style mappings.\vspace{-2mm}
  \item We present Shape-Matching GAN to transfer text styles, with a scale-controllable module designed to allow for adjusting the stylistic degree of the glyph with a continuous parameter as user input and generating diversified artistic text in real-time.
\end{itemize}

\section{Related Work}

\textbf{Image style transfer}.
Leveraging the powerful representation ability of neural networks, Gatys~\textit{et al.} pioneered on the Neural Style Transfer~\cite{gatys2016image}, where the style was effectively formulated as the Gram matrix~\cite{gatys2015texture} of deep features. Johnson~\textit{et al.} trained a feed-forward StyleNet~\cite{Johnson2016Perceptual} using the loss of Neural Style Transfer~\cite{gatys2016image} for fast style transfer~\cite{Wang2016Multimodal,huang2017adain,WCT-NIPS-2017,Li2017Diversified,Chen2017StyleBank}.
In parallel, Li~\textit{et al.}~\cite{Li2016Combining,Li2016Precomputed} represented styles by neural patches, which can better preserve structures for photo-realistic styles.
Meanwhile, other researchers regard style transfer as an image-to-image translation problem~\cite{Isola2017Image,Zhu2017Unpaired}, and exploited Generative Adversarial Network~(GAN)~\cite{goodfellow2014generative} to transfer specialized styles such as cartoons~\cite{chen2018cartoongan}, paintings~\cite{sanakoyeu2018style} and makeups~\cite{Choi2017StarGAN,chang2018pairedcyclegan}.
Compared to Gram-based and patch-based methods, GAN learns the style representation directly from the data, which can potentially yield more artistically rich results.

\textbf{Artistic text style transfer}.
The problem of artistic text style transfer was first raised by Yang~\textit{et al.}~\cite{Yang2017Awesome}.
The authors represented the text style using image patches, 
which suffered from a heavy computational burden due to the patch matching procedure. Driven by the progress of neural network, Azadi~\textit{et al.}~\cite{Azadi2017Multi} trained an MC-GAN for fast text style transfer, which, however, can only render $26$ capital letters.
Yang~\textit{et al.}~\cite{Yang2019TETGAN} recently collected a large dataset of text effects to train the network to transfer text effects for any glyph.
Unlike the aforementioned methods that assume the input style to be well-designed text effects,
a patch-based model UT-Effect~\cite{Yang2018Context} stylized the text with arbitrary textures and achieved glyph deformations by shape synthesis~\cite{rosenberger2009layered}, which shows promise for more application scenarios.
Compared to UT-Effect~\cite{Yang2018Context}, our GAN-based method further enables the continuous adjustment of glyph deformations via a controllable parameter in real-time.

\begin{figure}[t]
\centering
\includegraphics[width=1\linewidth]{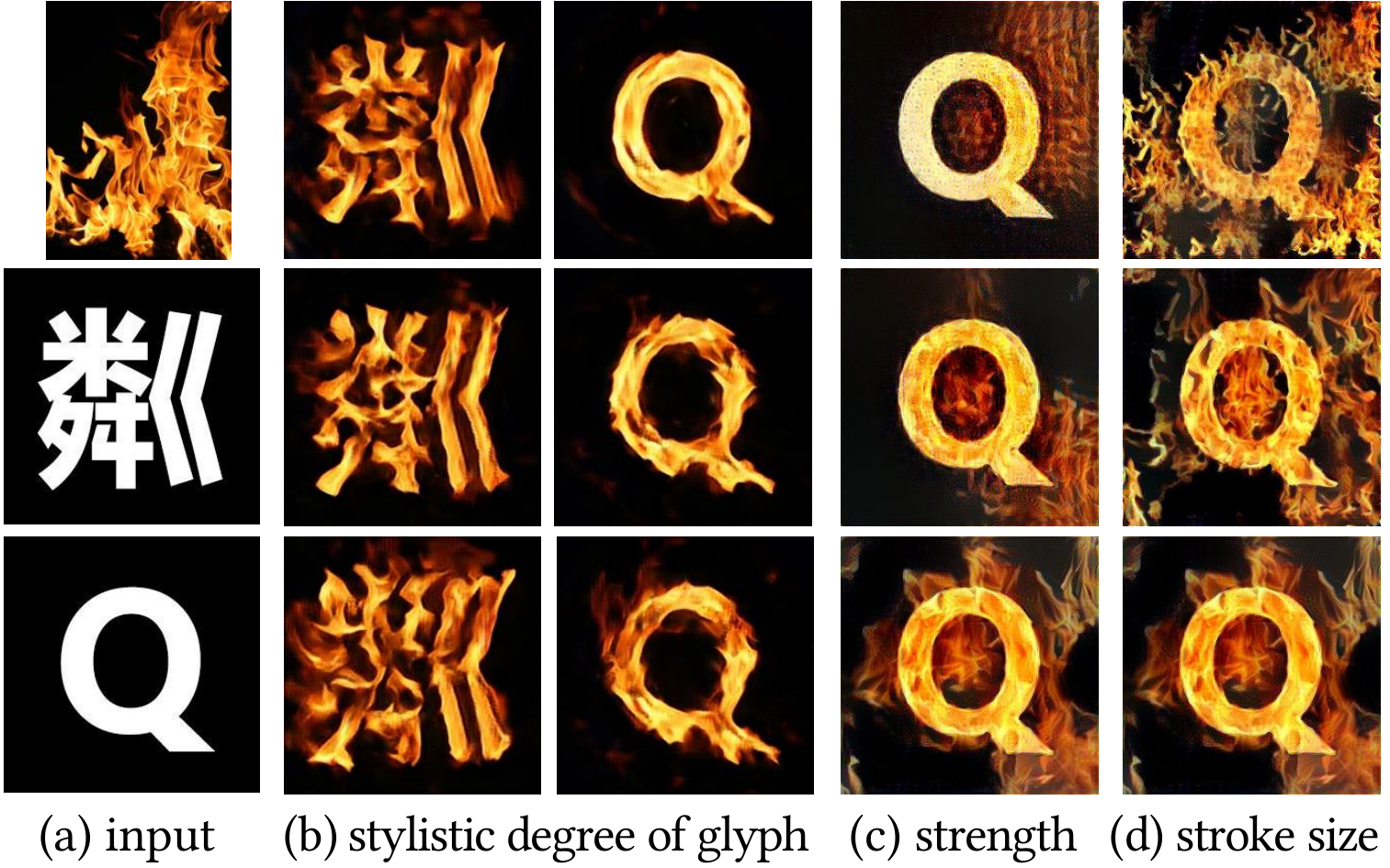}\vspace{-2mm}
\caption{Comparing different scale effects in text style transfer.
(a) shows the reference style and target text.
In remaining columns, each shows the results with increasing
(b) glyph deformation degree; (c) texture strength; and (d) stroke size. Results in (b) are generated by our proposed method, while (c) and (d) are generated by Neural Style Transfer~\cite{gatys2016image}.}\vspace{-3mm}
\label{fig:ana-ana}
\end{figure}

\textbf{Multi-scale style control}.
To the best of our knowledge, the research on the multi-scale style control currently focuses on two kinds of scales: the \textit{strength} and the \textit{stoke size} of the texture.
The texture strength determines the texture similarity between the result and the style image (Fig.~\ref{fig:ana-ana}(c)).
It is mainly controlled by a hyper-parameter to balance the content loss and style loss~\cite{gatys2016image}.
As a result, one has to re-train the model for different texture strengths.
Babaeizadeh~\textit{et al.}~\cite{Babaeizadeh2018AdjustableRS} performed efficient adjustment of the texture strength, with an auxiliary network to input additional parameters to modulate the style transfer process. 
Meanwhile, the stroke size depicts the scale of texture patterns (Fig.~\ref{fig:ana-ana}(d)), \textit{e.g.}, size or spatial frequency.
Jing~\textit{et al.}~\cite{jing2018stroke} 
proposed a stroke-controllable neural style transfer network (SC-NST) with adaptive receptive fields for stroke size control.
Our work explores the glyph deformation degree~(Fig.~\ref{fig:ana-ana}(b)), a different and important dimension of ``scale'' that is unexplored in prior work.

\section{Problem Overview}
\label{sec:analysis}


We start by giving operative requirements for our new task. Considering the \textit{maple} style for instance, it will look weird to have artistic text with the texture of leaves, but without the leave-like shapes (see an example in Fig.~\ref{fig:ablation3}), demonstrating the need of shape deformation and matching in addition to merely transferring texture patterns.
Meanwhile, the optimal scale to balance the legibility and artistry can vary a lot for different styles and text contents, not to mention the subjective variation among people. Taking Fig.~\ref{fig:ana-ana}(b) for example, one may see that the glyph with more complex strokes is more vulnerable to large glyph deformation~\cite{wang2015deepfont}. Therefore, users will enjoy the freedom to navigate through the possible scale space of glyph deformations, without the hassle of re-training one model per scale. Concretely, a controllable artistic text style transfer shall ensure:
\begin{itemize}\vspace{-2mm}
  \item \textit{Artistry}: The stylized text should mimic the shape characteristics of the style reference, at any scale.\vspace{-2mm}
  \item \textit{Controllability}: The glyph deformation degree needs be adjusted in a quick and continuous way.\vspace{-2mm}
\end{itemize}
The two requirements distinguish our problem from those studied by previous multi-scale style transfer methods, which are either unable to adjust the shape at all~\cite{Babaeizadeh2018AdjustableRS,jing2018stroke} (\textit{e.g.},  Fig.~\ref{fig:ana-ana}(c)(d)) or fail to do so efficiently~\cite{Yang2018Context}.

\begin{figure}[t]
\centering
\includegraphics[width=1\linewidth]{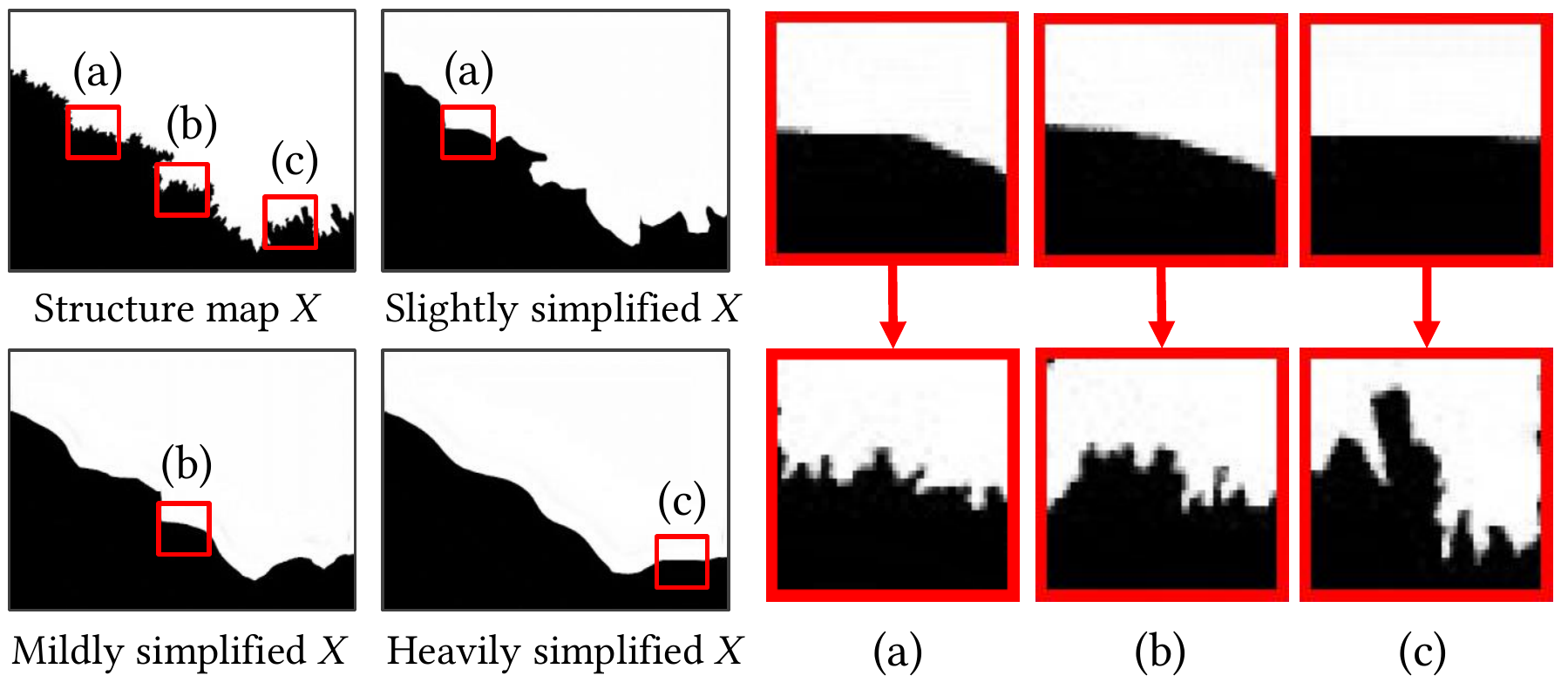}\vspace{-1mm}
\caption{Illustration of bidirectional shape matching. Left two columns: a leaf-shaped structure map and its three backward simplified versions. Right columns: forward shape mappings under (a) slight, (b) moderate, and (c) heavy deformations.}\vspace{-4mm}
\label{fig:ana-map}
\end{figure}

Our solution to this problem is a novel bidirectional shape matching strategy. As illustrated in Fig.~\ref{fig:ana-map}, the target structure map is first (\underline{backward}) simplified to different coarse levels, and then its stylish shape features can be characterized by the (\underline{forward}) multi-level coarse-to-fine shape mappings, to realize multi-scale transfer. As shown in Fig.~\ref{fig:ana-map}(a)-(c), similar horizontal strokes at different levels are mapped to different shapes, and the coarser the level, the greater the deformation between the mapped shapes.
\textit{Artistry} is met since the targets in these mappings are exactly the reference fine-level stylish shapes while \textit{Controllability} can be achieved via training a feed-forward network.


To sum up, we formulate the new task of scale-controllable artistic text style transfer as learning \textit{the function to map the style image from different coarse levels back to itself in a fast feed-forward way}. Still, two technical roadblocks remain to be cleared. First, how to simplify the shape to make the obtained mapping applicable to the text images.
Second, how to learn the many-to-one (multiple coarse levels to a fine level) mapping without model collapse.
Sec.~\ref{sec4} will detail how we address these challenges with our network design.

\begin{figure*}[t]
  \centering
    \includegraphics[width=0.98\linewidth]{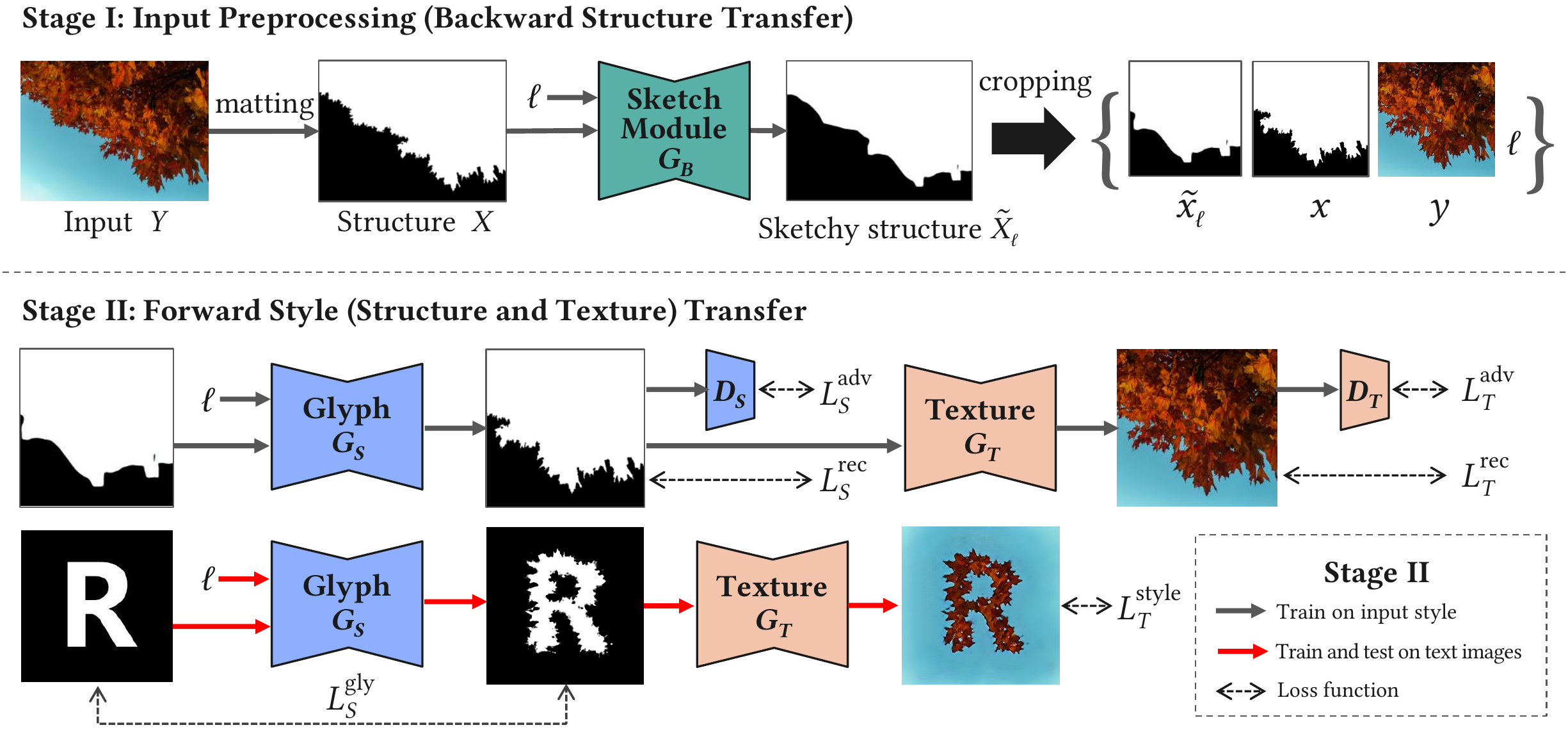}\vspace{-2mm}
  \caption{Overview of our bidirectional shape matching framework.\vspace{-3mm}}
  \label{fig:overview}
\end{figure*}

\section{Shape-Matching GAN}
\label{sec4}
\vspace{-1mm}


Assume that $Y$ and $I$ denote the style image and text image provided by users, respectively.
We study the problem of designing a feed-forward stylization model $G$ to render artistic text under different deformation degrees controlled by a parameter $\ell\in[0,1]$, where larger $\ell$ corresponds to greater deformations.
We further decompose the style transfer process into two successive stages: structure transfer and texture transfer, which are modelled by generators $G_S$ and $G_T$, separately. The advantage of such decomposition is that we can disentangle the influence of textures and first focus on the key shape deformation problem. We denote $G=G_T\circ G_S$, and formulate the stylization process as:\vspace{-0.5mm}
\begin{equation}
  I_\ell^Y=G_T(G_S(I,\ell)), ~~I_\ell^Y\sim p(I_\ell^Y|I,Y,\ell),\vspace{-0.5mm}
\end{equation}
where the target statistic $p(I_\ell^Y)$ of the stylized image $I_\ell^Y$ is characterized by the text image $I$, the style image $Y$ and the controllable parameter $\ell$.

As outlined in Sec.~\ref{sec:analysis}, our solution to structure transfer is bidirectional shape matching.						
Assume the structure map $X$ to mask the shape of the style subject in $Y$ is given, which can be easily obtained by image editing tools such as Photoshop or existing image matting algorithms.
In the stage of backward structure transfer, we preprocess $X$ to obtain training pairs $\{\tilde{X}_\ell,X\}$ for $G_S$, where $\tilde{X}_\ell$ is a sketchy (coarse) version of $X$ with the shape characteristics of the text, and $\ell$ controls the coarse level.
In the stage of forward structure transfer, $G_S$ learns from $\{\tilde{X}_\ell,X\}$ to stylize the glyph with various deformation degrees. Fig.~\ref{fig:overview} summarizes the overall framework built upon two main components:\vspace{-2mm}
\begin{itemize}
  \item \textbf{Glyph Network} $G_S$: It learns to map $\tilde{X}_\ell$ with deformation degree $\ell$ to $X$ during training. In testing, it transfers the shape style of $X$ onto the target text image $I$, producing the structure transfer result $I_\ell^X$.\vspace{-2mm}
  \item \textbf{Texture Network} $G_T$: It renders the texture in the style image $Y$ on $I_\ell^X$ to yield the final artistic text $I_\ell^Y$.\vspace{-2mm}
\end{itemize}
The  generators are accompanied with corresponding discriminators $D_S$ and $D_T$ to improve the quality of the results through adversarial learning. 
In the following, we present the details of our bidirectional shape matching and the proposed controllable module that enables $G_S$ to learn multi-scale glyph deformations in Sec.~\ref{sec:glyph_network}. The texture transfer network $G_T$ is then introduced in Sec.~\ref{sec:texture_network}.

\begin{figure}[t]
  \centering
    \includegraphics[width=\linewidth]{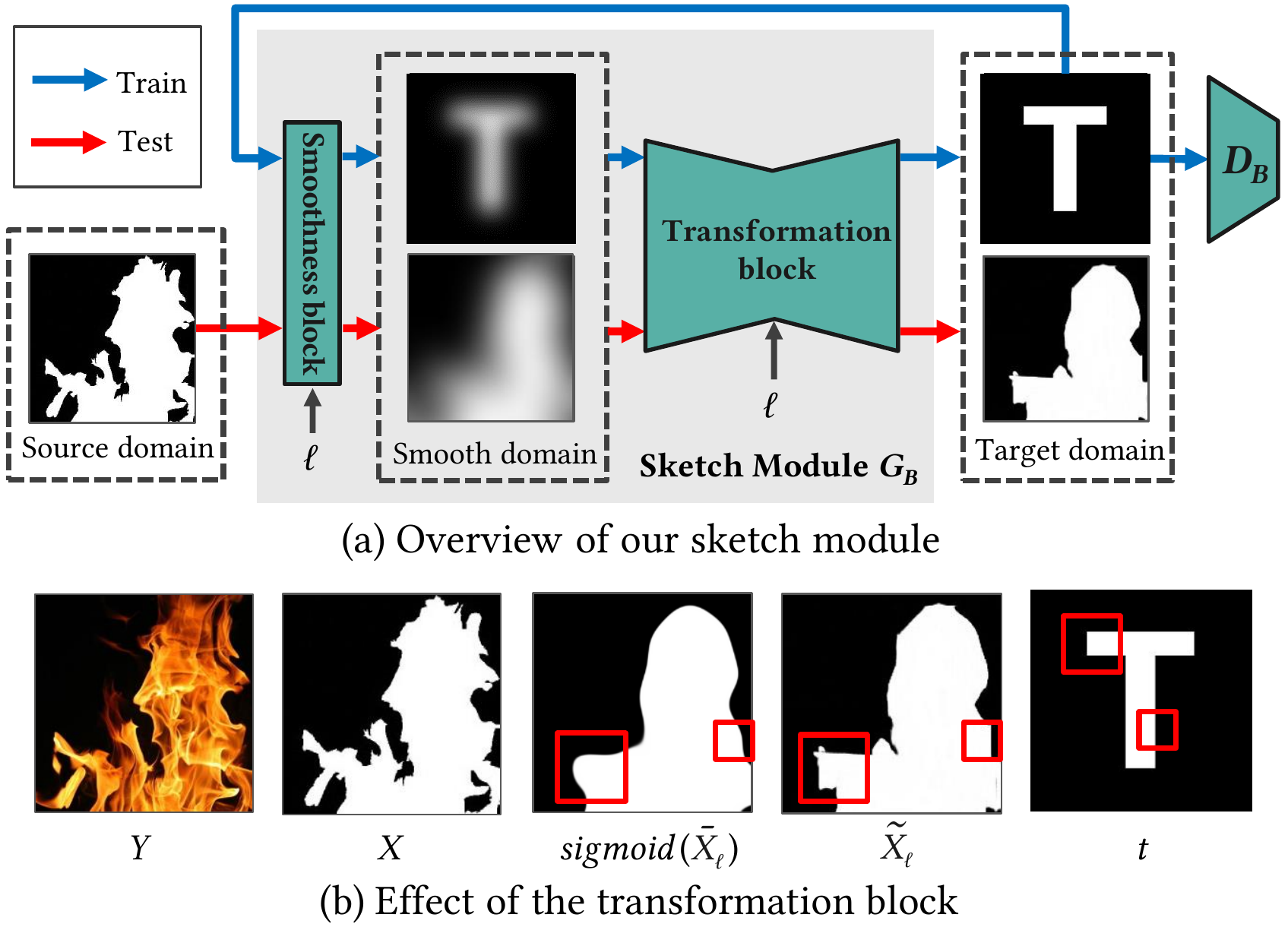}\vspace{-1mm}
  \caption{Backward structure transfer by sketch module~$G_B$}\vspace{-5mm}
  \label{fig:deblur}
\end{figure}

\subsection{Bidirectional Structure Transfer ($G_S$)}
\label{sec:glyph_network}



\textbf{Backward structure transfer.}
To transfer the glyph characteristics to $X$ at different coarse levels, we propose a sketch module $G_B$ composed of a smoothness block and a transformation block, as shown in Fig.~\ref{fig:deblur}(a).
Inspired by the Gaussian scale-space representation~\cite{babaud1986uniqueness,perona1990scale} for simplifying images at different scales, our smoothness block is set as a fixed convolutional layer with Gaussian kernel, whose standard deviation $\sigma=f(\ell)$ is controlled by $\ell$ and a linear function $f(\cdot)$.
Our key idea is to bridge the source style domain and the target text domain using the smoothness block that maps the text image and $X$ into a smooth domain, where the details are eliminated and the contours demonstrate similar smoothness. Structure transfer is then achieved by training the transformation block to map the smoothed text images back to the text domain to learn the glyph characteristics. The advantages of our sketch module are twofold: 1) the coarse level (and thus the deformation degree) can be naturally parameterized by $\sigma$; and 2) the training process of $G_B$ only requires easily accessible text images. Once trained, it can be applied to arbitrary input styles.



For training $G_B$, we sample a text image $t$ from the text dataset provided by~\cite{Yang2019TETGAN} and a parameter value $\ell$ from $[0,1]$. $G_B$ is tasked to reconstruct $t$:
\begin{equation}\label{eq:deblur_rec_loss}
  \mathcal{L}_B^{\text{rec}}=\mathbb{E}_{t,\ell}[\|G_B(t,\ell)-t\|_{1}].
\end{equation}
In addition, we impose a conditional adversarial loss to force $G_B$ to generate more text-like contours:
\begin{equation}\label{eq:deblur_adv_loss}
\begin{aligned}
  \mathcal{L}_B^{\text{adv}}=&~\mathbb{E}_{t,\ell}[\log D_B(t,\ell,\bar{t}_\ell)]\\
  +&~\mathbb{E}_{t,\ell}[\log(1-D_B(G_B(t,\ell),\ell,\bar{t}_\ell))],
\end{aligned}
\end{equation}
where $D_B$ learns to determine the authenticity of the input image and whether it matches the given smoothed image $\bar{t}_\ell$ and the parameter $\ell$. Thus, the total loss takes the form of
\begin{equation}\label{eq:deblur_total_loss}
  \min_{G_B}\max_{D_B}\lambda_B^{\text{adv}}\mathcal{L}_B^{\text{adv}}+\lambda_B^{\text{rec}}\mathcal{L}_B^{\text{rec}}.
\end{equation}

Finally, by applying trained $G_B$ to $X$ with various level~$\ell$, we can obtain the corresponding sketchy shape $\tilde{X}_\ell=G_B(X,\ell)$. An example is shown in Fig.~\ref{fig:deblur}(b). The simply thresholded Gaussian representation $sigmoid(\bar{X}_\ell)$ (by replacing the transformation block with a sigmoid layer) does not match the shape of the text. In contrast, our sketch module effectively simplifies the flame profile to the shape of  strokes in the red box regions, thus providing a more robust shape mapping for the glyph network.

\textbf{Forward structure transfer.} Having obtained $\{\tilde{X}_\ell\}$, $\ell\in[0,1]$, we now train the glyph network $G_S$ to map them to the original $X$ so that $G_S$ can characterize the shape features of $X$ and transfer these features to the target text. Note that our task is a many-to-one mapping, and we only have a single example $X$.
The network should be carefully designed to avoid just memorizing the ground truth $X$ and falling into model collapse, namely, yielding very similar results regardless of the parameter $\ell$ during testing.

\begin{figure}[t]
  \centering
    \includegraphics[width=0.84\linewidth]{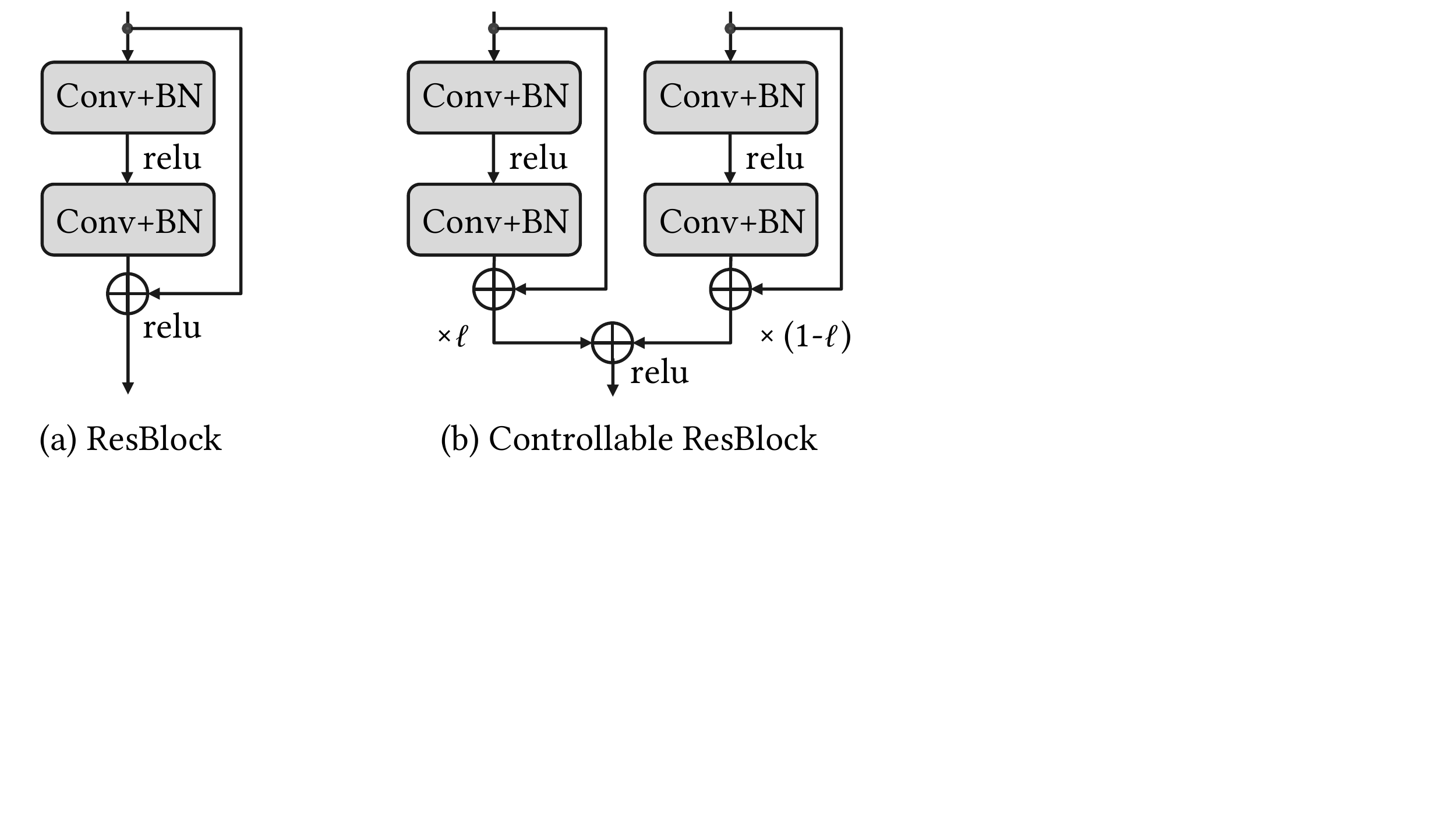}\vspace{-1mm}
  \caption{Controllable ResBlock}\vspace{-3mm}
  \label{fig:resblock}
\end{figure}

To tackle this challenging task, we employ two strategies: \textit{data augmentation} and \textit{Controllable ResBlock}.
First, $X$ and $\tilde{X}_\ell$ are randomly cropped into sub-image pairs $\{x,\tilde{x}_\ell\}$ to gather as a training set.
Second, we build $G_S$ upon the architecture of StyleNet~\cite{Johnson2016Perceptual}, and propose a very simple yet effective Controllable ResBlock to replace the original ResBlock~\cite{he2016deep} in the middle layers of StyleNet. Our Controllable ResBlock is a linear combination of two ResBlocks weighted by $\ell$, as shown in Fig.~\ref{fig:resblock}. For $\ell=1~(0)$, $G_S$ degrades into the original StyleNet, and is solely tasked with the greatest (tiniest) shape deformation to avoid the many-to-one problem. Meanwhile for $\ell\in(0,1)$, $G_S$ tries to compromise between the two extremes.

In terms of the loss, $G_S$ aims to approach the ground truth $X$ in an $L_1$ sense and confuse the discriminator $D_S$:
\begin{equation}\label{eq:structure_rec_loss}
  \mathcal{L}_S^{\text{rec}}=\mathbb{E}_{x,\ell}[\|G_S(\tilde{x}_\ell,\ell)-x\|_{1}],
\end{equation}
\begin{equation}\label{eq:structure_adv_loss}
\begin{aligned}
  \mathcal{L}_S^{\text{adv}}=&~\mathbb{E}_{x}[\log D_S(x)]\\
  +&~\mathbb{E}_{x,\ell}[\log(1-D_S(G_S(\tilde{x}_\ell,\ell)))].
\end{aligned}
\end{equation}
For some styles with large $\ell$, the text $t$ could be too severely deformed to be recognized. Thus we propose an optional glyph legibility loss to force the structure transfer result $G_S(t,\ell)$ to maintain the main stroke part of $t$:
\begin{equation}\label{eq:structure_gly_loss}
  \mathcal{L}_S^{\text{gly}}=\mathbb{E}_{t,\ell}[\|(G_S(t,\ell)-t)\otimes M(t)\|_{1}],
\end{equation}
where $\otimes$ is the element-wise multiplication operator, and $M(t)$ is a weighting map based on distance field whose pixel value increases with its distance to the nearest text contour point of $t$.
The overall loss for $G_S$ is as follows:
\begin{equation}\label{eq:structure_total_loss}
  \min_{G_S}\max_{D_S}\lambda_S^{\text{adv}}\mathcal{L}_S^{\text{adv}}+\lambda_S^{\text{rec}}\mathcal{L}_S^{\text{rec}}
  +\lambda_S^{\text{gly}}\mathcal{L}_S^{\text{gly}}.
\end{equation}

\begin{figure*}[t]
\centering
\includegraphics[width=1\linewidth]{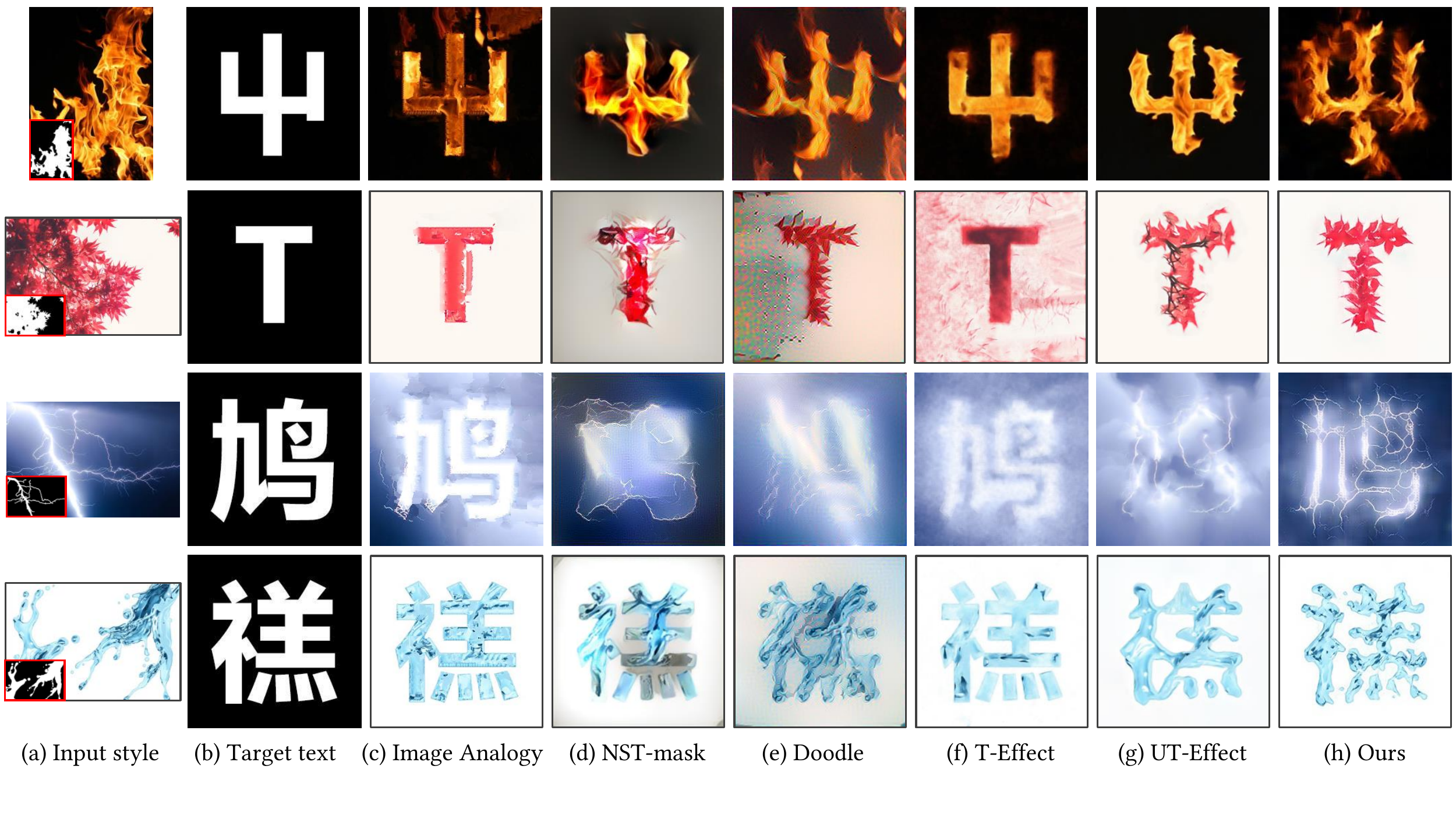}\vspace{-2mm}
\caption{Comparison with state-of-the-art methods on various styles. (a) Input style with its structure map in the lower-left corner. (b) Target text. (c) Image Analogy~\cite{Hertzmann2001Image}. (d) Neural Style Transfer~\cite{gatys2016image} with spatial control~\cite{Gatys2017Controlling}.  (e) Neural Doodle~\cite{Champandard2016Semantic}. (f) T-Effect~\cite{Yang2017Awesome}. (g) UT-Effect~\cite{Yang2018Context}. (h) Our style transfer results. We manually select the suitable deformation degrees for UT-Effect~\cite{Yang2018Context} and out method.}
\vspace{-2mm}
\label{fig:comparison}
\end{figure*}

\subsection{Texture Transfer ($G_T$)}
\label{sec:texture_network}


Given the structure transfer result $I^X_\ell=G_S(I,\ell)$, the texture rendering task can be formulated as a standard image analogy problem such that $X:Y::I^X_\ell:I^Y_\ell$~\cite{Hertzmann2001Image}, which can be well solved by existing algorithms like the greedy-based Image Analogy~\cite{Hertzmann2001Image} and the optimization-based Neural Doodle~\cite{Champandard2016Semantic}. To build an end-to-end fast text stylization model, we instead train a feed-forward network $G_T$ for texture rendering. Similar as in training $G_S$, we first use random cropping to obtain adequate training pairs $\{x,y\}$ from $X$ and $Y$. Then we train $G_T$ using the reconstruction loss and conditional adversarial loss:
\begin{equation}\label{eq:texture_rec_loss}
  \mathcal{L}_T^{\text{rec}}=\mathbb{E}_{x,y}[\|G_T(x)-y\|_{1}],
\end{equation}
\begin{equation}\label{eq:texture_adv_loss}
\begin{aligned}
  \mathcal{L}_T^{\text{adv}}=&~\mathbb{E}_{x,y}[\log D_T(x,y)]\\
  +&~\mathbb{E}_{x,y}[\log(1-D_T(x,G_T(x)))].
\end{aligned}
\end{equation}
The overall style rendering performance on the sampled text image $t$ is further taken into account 
by adding the style loss $\mathcal{L}_T^{\text{style}}$ proposed in Neural Style Transfer~\cite{gatys2016image}.
Finally, the objective of texture transfer can be defined as:
\begin{equation}\label{eq:texture_total_loss}
  \min_{G_T}\max_{D_T}\lambda_T^{\text{adv}}\mathcal{L}_T^{\text{adv}}+\lambda_T^{\text{rec}}\mathcal{L}_T^{\text{rec}}
  +\lambda_T^{\text{style}}\mathcal{L}_T^{\text{style}}.
\end{equation}

\section{Experimental Results}

\subsection{Implementation Details}
\vspace{-1mm}

\textbf{Network architecture}. We adapt our generators from the Encoder-Decoder architecture of StyleNet~\cite{Johnson2016Perceptual} with six ResBlocks, except that $G_S$ uses the proposed Controllable ResBlock instead.
Our discriminators follow PatchGAN~\cite{Isola2017Image}.
To prevent over-fitting, we apply dropout~\cite{srivastava2014dropout} with a rate of $0.5$ to the residual blocks.
Since the structure map contains many saturated areas,
we add gaussian noises onto the input of $G_S$ and $G_T$ to avoid ambiguous problem.
It also empowers our network to generate diversified results during testing as shown in Fig.~\ref{fig:teaser}(d) (animation).
Code and pretrained models are available at: \url{https://github.com/TAMU-VITA/ShapeMatchingGAN}.

\textbf{Network training}. We randomly crop the style image to $256\times256$ sub-images for training.
The Adam optimizer is adopted with a fixed learning rate of $0.0002$.
To stabilize the training of $G_S$, we gradually increase the sampling range of $\ell$.
Specifically, $G_S$ is first trained with a fixed $\ell=1$ to learn the greatest deformation.
Then we copy the parameters from the trained half part in Controllable ResBlocks to the other half part and use
$\ell\in\{0,1\}$ to learn two extremes. Finally, $G_S$ is tuned on $\ell\in\{i/K\}_{i=0,...,K}$. We find that $K=3$ is sufficient for $G_S$ to infer the remaining intermediate scales.
The linear function to control the standard deviation of the Gaussian kernel is $f(\ell)=16\ell+8$.
For all experiments, we set
$\lambda^{\text{rec}}_B=\lambda^{\text{rec}}_S=\lambda^{\text{rec}}_T=100$,
$\lambda^{\text{adv}}_B=\lambda^{\text{adv}}_T=1$,
$\lambda^{\text{adv}}_S=0.1$, and $\lambda^{\text{style}}_T=0.01$.

\begin{figure*}[t]
\centering
\includegraphics[width=1\linewidth]{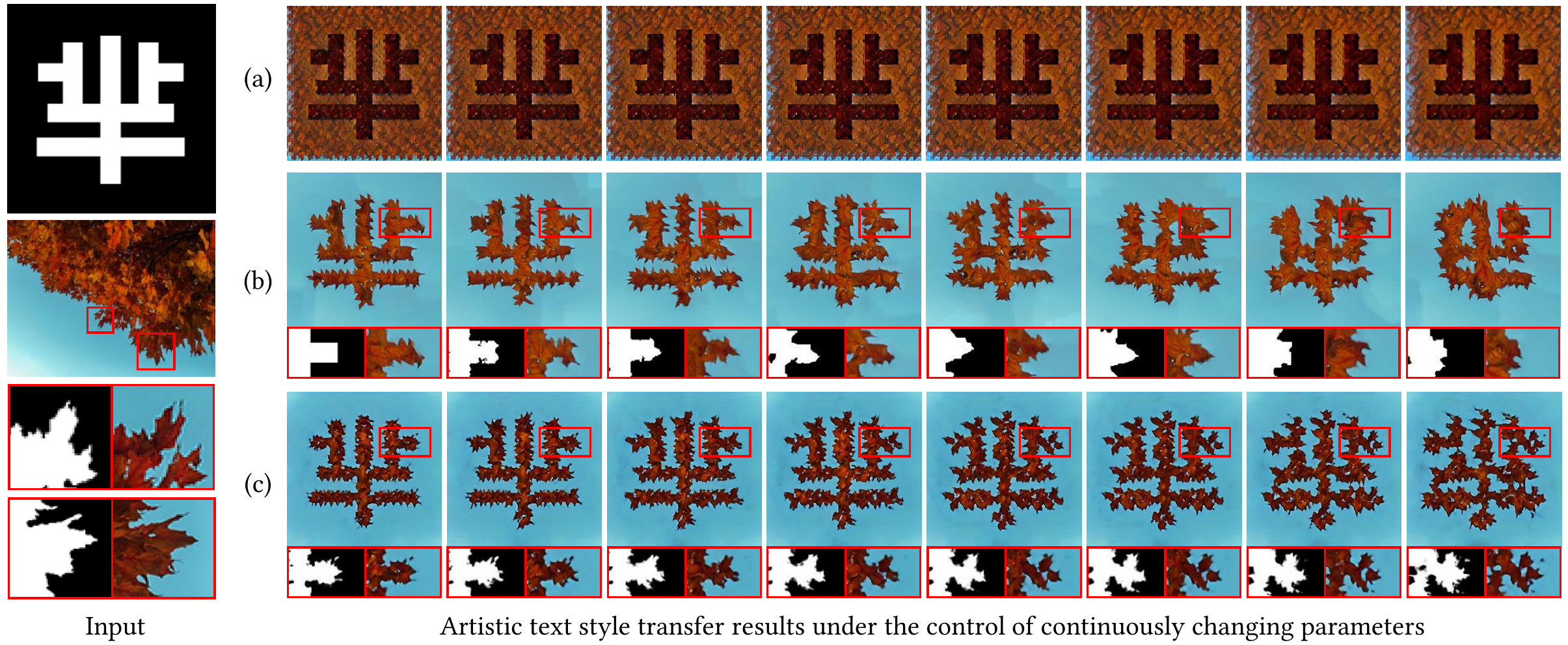}\vspace{-1mm}
\caption{Qualitative comparison between the proposed method and other scale-controllable style transfer methods. For the first column, from top to bottom: target text, style image, the enlarged patches from the style image and their corresponding structure maps. Remaining columns: Results by  (a) stroke-controllable neural style transfer (SC-NST)~\cite{jing2018stroke} with stroke size evenly increasing from $256$ to $768$; (b) UT-Effect~\cite{Yang2018Context} with resolution level evenly increasing from $1$ to $7$; (c) the proposed method with $\ell$ evenly increasing from $0$ to $1$. All results are produced by one single model for each method. For UT-Effect~\cite{Yang2018Context} and our method, the red box region is shown enlarged in the bottom with the corresponding structure map provided for better visual comparison.}\vspace{-2mm}
\label{fig:comparison2}
\end{figure*}

\subsection{Comparisons with State-of-the-Art Methods}

\textbf{Artistic text style transfer}. In Fig~\ref{fig:comparison}, we present the qualitative comparison with five state-of-the-art style transfer methods:
Image Analogy~\cite{Hertzmann2001Image}, NST~\cite{Gatys2017Controlling}, Doodle~\cite{Champandard2016Semantic}, T-Effect~\cite{Yang2017Awesome} and UT-Effect~\cite{Yang2018Context}\footnote{For NST, We build upon its public model and implement the spatial control function introduced in~\cite{Gatys2017Controlling}. Meanwhile, results of other methods are obtained by public models or provided by the authors.}.
These methods are selected because they are all one-shot supervised (or can be adapted to a supervised version) for a fair comparison, which transfer styles with a single style image and its structure map.

Image Analogy~\cite{Hertzmann2001Image} and T-Effect~\cite{Yang2017Awesome} directly copy the texture patches to the text region, yielding rigid and unnatural contours.
NST~\cite{Gatys2017Controlling} and Doodle~\cite{Champandard2016Semantic} are deep learning-based methods, where the
shape characteristics of the style are implicitly represented by deep features.
Thus these methods could modify the glyph contours but often lead to unrecognizable text.
In terms of texture, they suffer from color deviations and checkerboard artifacts.
UT-Effect~\cite{Yang2018Context} explicitly matches the glyph to the style at a patch level. However, image patches are not always robust. For example, in the \textit{maple} style, the leaf shapes are not fully transferred to the vertical stroke.
In addition, texture details are lost due to the patch blending procedure. By comparison, our network is able to learn accurate shape characteristics through the proposed bidirectional shape matching strategy, and transfers vivid textures via adversarial learning, which together leads to the most visually appealing results.

To quantitatively evaluate the performance of the compared methods, we conducted a user study on the Amazon Mechanical Turk platform where observers were given images pairs and asked to select which one is of the best style similarity with the reference style image while maintaining legibility. A total of $18$ styles are used and for each style $15$ image pairs were rated by $10$ observers, obtaining $2,700$ selection results. The proposed method obtains the best average preference ratio of $0.802$, while the average preference ratios of Analogy~\cite{Hertzmann2001Image}, NST-mask~\cite{gatys2016image}, Doodle~\cite{Champandard2016Semantic}, T-Effect~\cite{Yang2017Awesome} and UT-Effect~\cite{Yang2018Context} are $0.513$, $0.376$, $0.537$, $0.230$, $0.542$, respectively.
This user study shows that our method is highly preferred by users, which quantitatively verifies the superiority of our method.

\textbf{Scale-controllable style transfer}. In Fig.~\ref{fig:comparison2}, we present the qualitative comparison with two scale-controllable style transfer methods: SC-NST~\cite{jing2018stroke} and UT-Effect~\cite{Yang2018Context}.
SC-NST~\cite{jing2018stroke} does not synthesize the textures in the correct region due to its unsupervised setting. Regardless of this factor, it can adjust the texture size, but is ineffective in controlling the glyph deformation.
UT-Effect~\cite{Yang2018Context} matches boundary patches at multiple resolutions for structure transfer, which has several drawbacks:
First, as shown in Fig.~\ref{fig:comparison2}(b), the greedy-based patch matching fails to global consistently stylize the glyph. Second, the patch blending procedure inevitably eliminates many shape details. Third, the continuous transformation is not supported.
On the contrary, the proposed method achieves continuous transformation with fine details, showing a smooth growing process of the leaves as they turn more luxuriant.
In terms of efficiency, for $256\times256$ images in Fig.~\ref{fig:comparison2},
the released MATLAB-based UT-Effect~\cite{Yang2018Context} requires about 100 s per image with Intel Core i7-6500U CPU (no GPU version available). In comparison, our feed-forward method only takes about 0.43 s per image with Intel Xeon E5-2650 CPU and \textbf{16 ms per image} with a GeForce GTX 1080 Ti GPU, which implies a potential of nearly real-time user interaction.

\subsection{Ablation Study}
\label{sec:ablation_study}

\begin{figure}[t]
  \centering
  \includegraphics[width=0.98\linewidth]{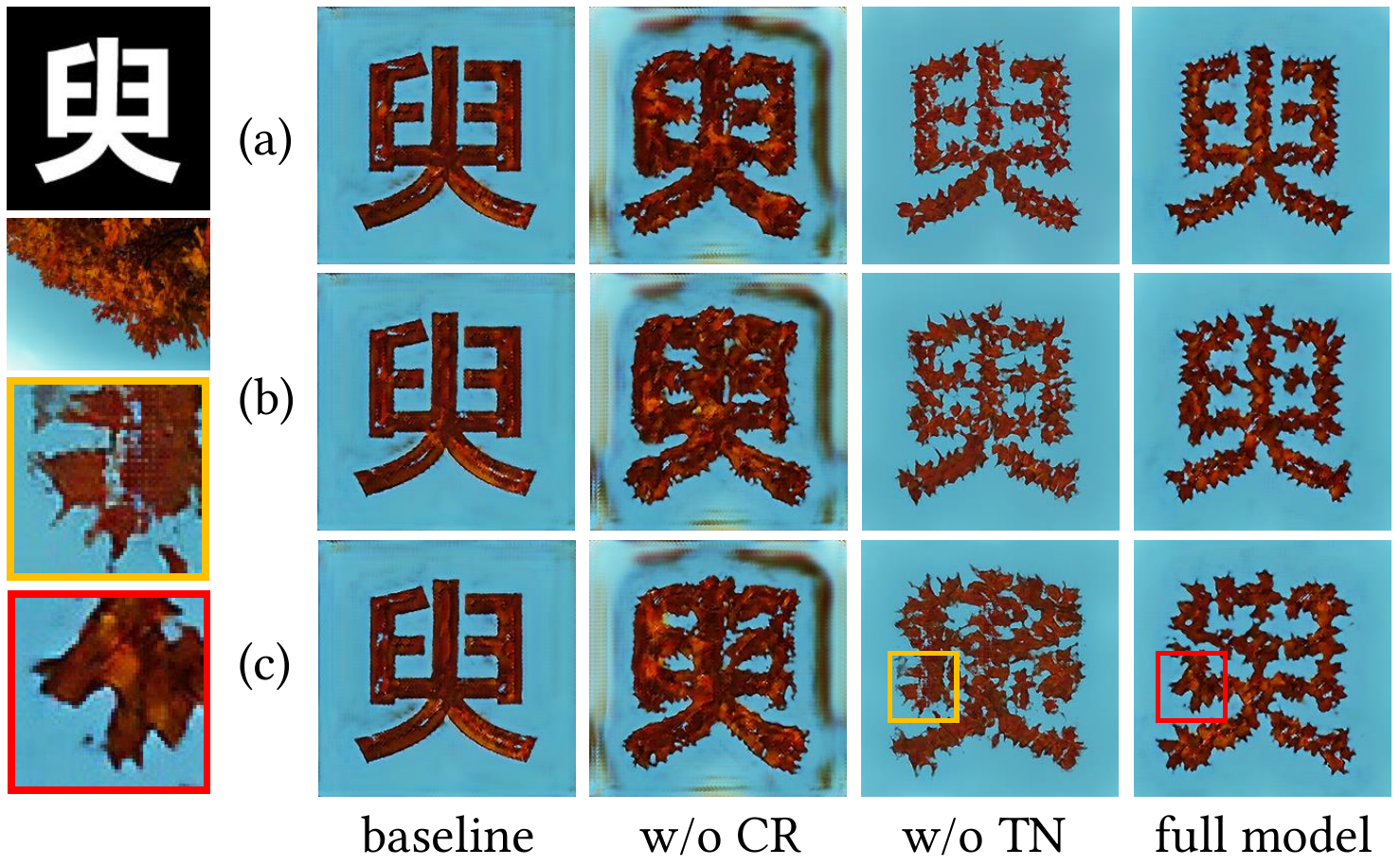}\vspace{-2mm}
  \caption{Analysis for the network configurations in controllable artistic text style transfer.
  For the first column, from top to bottom: target text, style image, the enlarged patches from the results without and with the texture network, respectively. Remaining columns: (a)-(c) Results with $\ell=0.0,0.5,1.0$, respectively.}\label{fig:ablation3}\vspace{-1mm}
\end{figure}

\textbf{Network architecture}. To analyze each component in our model, we design the following experiments with different configurations:\vspace{-3mm}
\begin{itemize}
\setlength{\itemsep}{0pt}
\setlength{\parsep}{0pt}
\setlength{\parskip}{0pt}
  \item Baseline: Our baseline model contains only a texture network trained to directly map the structure map $X$ back to the style image $Y$.
  \item W/o CR: This model contains a na\"{\i}ve glyph network and a texture network. The na\"{\i}ve glyph network is controlled by $\ell$ via the commonly used label concatenation instead of using the Controllable ResRlock (CR).
  \item W/o TN: This model contains a single glyph network without the Texture Network (TN), and is trained to directly map the sketchy structure map $\tilde{X}_\ell$ to $Y$.
  \item Full model: The proposed model with both the glyph network and the texture network.\vspace{-2mm}
\end{itemize}
Fig.~\ref{fig:ablation3} displays the stylization results of these models.
Without structure transfer, the contours of the stylized text by baseline model are rigid, showing poor shape consistency with the reference style.
The na\"{\i}ve glyph network could create leaf-like shapes, but fails to learn the challenging many-to-one mapping. It simply ignores the conditional $\ell$, and generates very similar results.
This problem is well solved by the proposed Controllable ResBlock. As shown in the fourth column of Fig.~\ref{fig:ablation3}, our glyph network can even learn the multi-scale structure transfer and texture transfer simultaneously, although the rendered texture is flat and has checkerboard artifacts.
By handing the texture transfer task over to a separate texture network, our full model can synthesize high-quality artistic text, with both shape and texture consistency w.r.t. the reference style.

\begin{figure}[t]
  \centering
  \includegraphics[width=0.98\linewidth]{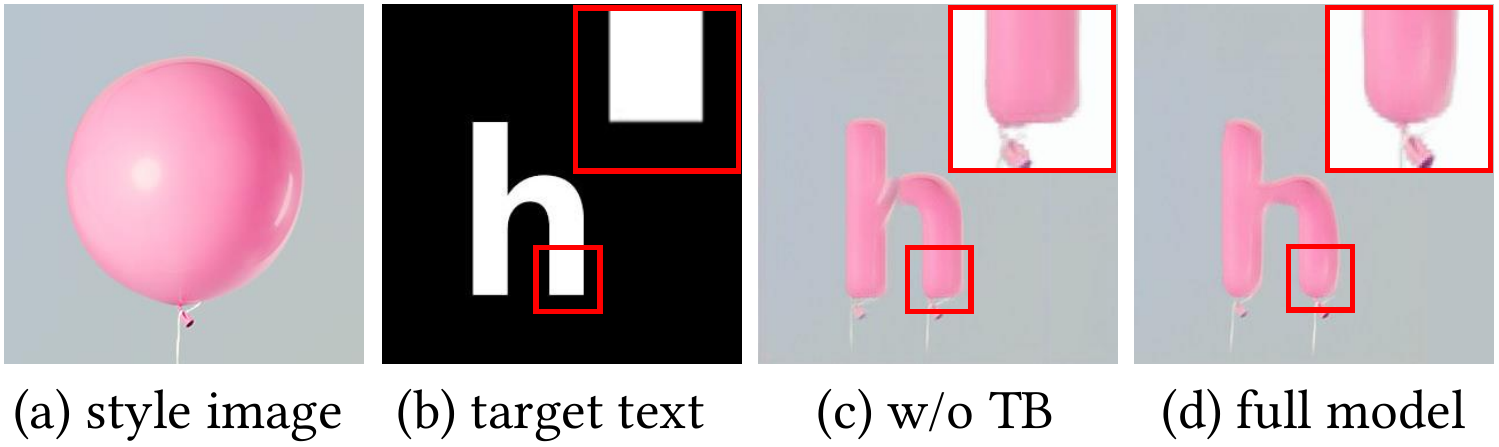}\vspace{-1mm}
  \caption{Effect of the proposed sketch module. The red box area is enlarged and the contrast is enhanced in the top right corner for better visual comparison.}\label{fig:ablation1}\vspace{-4mm}
\end{figure}

\textbf{Sketch module}. In Fig.~\ref{fig:ablation1}, we examine the effect of the sketch module $G_B$ through a comparative experiment.
As introduced in Sec.~\ref{sec:glyph_network}, our sketch module aims to transfer the shape characteristics of the text to the style image to provide a robust mapping between the source and target domains.
To make a comparison, we replace the Transformation Block (TB) in $G_B$ with a simple sigmoid layer. The resulting na\"{\i}ve sketch module is still able to simplify the shape, but cannot match it with the glyph.
Without robust mappings, the shape of the stylized text is not correctly adjusted and is as rigid as the input text as shown in Fig.~\ref{fig:ablation1}(c). By contrast, our full model successfully synthesizes a rounded h-shaped balloon in Fig.~\ref{fig:ablation1}(d).

\begin{figure}[t]
  \centering
  \includegraphics[width=0.98\linewidth]{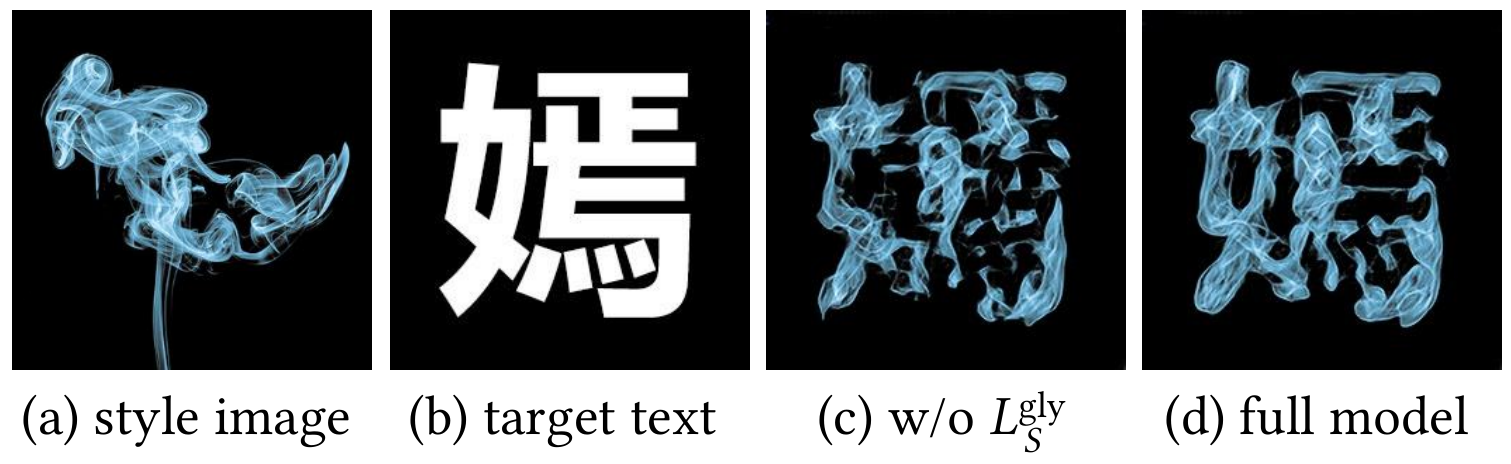}\vspace{-2mm}
  \caption{Effect of the glyph legibility loss $\mathcal{L}_S^{\text{gly}}$.}\label{fig:ablation2}\vspace{-2mm}
\end{figure}

\textbf{Loss function}. We study the effect of the glyph legibility loss (Eq.~(\ref{eq:structure_gly_loss})) in Fig.~\ref{fig:ablation2}.
When transferring a trickle of wafting smoke onto a rigid Chinese character with a high deformation degree $\ell=0.75$, the strokes of this character demonstrate irregular shapes, uneven thickness, and even fractures in Fig.~\ref{fig:ablation2}(c). Although very similar to the style, the character is unrecognizable.
As shown in Fig.~\ref{fig:ablation2}(d), by setting $\lambda^{\text{gly}}_S=1$, our glyph legibility loss effectively preserves the trunk of the strokes, while allowing a high freedom to deform the contours of the strokes, thus achieving a balance between legibility and artistry.

\begin{figure}[t]
  \centering
  \includegraphics[width=0.8\linewidth]{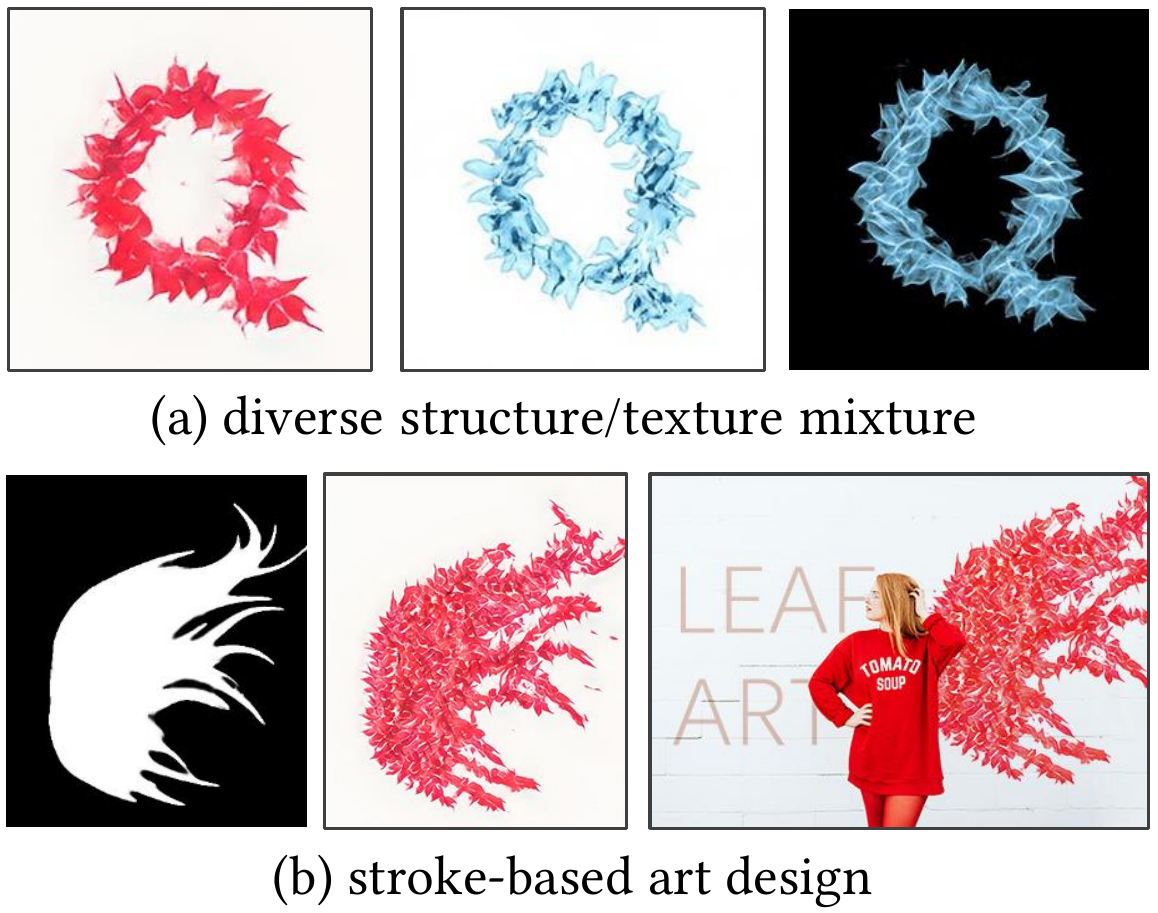}\vspace{-2mm}
  \caption{Applications of our method.}\label{fig:app}\vspace{-5mm}
\end{figure}

\subsection{Applications}

In addition to the poster and dynamic typography design shown in Fig.~\ref{fig:teaser}(d)-(f), we further present two other applications of our method as follows.

\textbf{Structure/texture mash-up}. The disentanglement of structures and textures enables us to combine different styles to create some brand-new text styles. Some examples are shown in Fig.~\ref{fig:app}(a), where we apply the textures of \textit{maple}, \textit{water} and \textit{smoke} to the text with the shape characteristics of \textit{maple}, respectively.

\textbf{Stroke-based art design}. Since no step specially tailored for the text is used, our method can be
easily extended to style transfer on more general shapes such as symbols and icons. In Fig.~\ref{fig:app}(b), we show an example for synthesizing wings made of maple leaves from a user-provided icon.

\section{Conclusion}
\vspace{-1mm}
In this paper, we present a fast artistic text style transfer deep network that allows for flexible, continuous control of the stylistic degree of the glyph. We formulate the task of glyph deformation as a coarse-to-fine mapping problem and propose a bidirectional shape matching framework. A novel sketch module is proposed to reduce the structural discrepancy between the glyph and style to provide robust mappings. Exploiting the proposed Controllable ResBlock, our network is able to effectively learn the many-to-one shape mapping for multi-scale style transfer.
We validate the effectiveness and robustness of our method by comparisons with state-of-the-art style transfer algorithms.
In future work, we would like to explore a smoothness block adaptable to different styles in place of the fixed Gaussian filter, and extend the model to artistic text video synthesis.

\section{Acknowledgement}
This work was supported in part by National Natural Science Foundation of China under contract No.~61772043, and in part by Beijing Natural Science Foundation under contract No.~L182002 and No.~4192025. This work was supported by China Scholarship Council.~We thank the Unsplash users (Aaron Burden, Andre Benz, Brandon Morgan, Brooke Cagle, Florian Klauer, Grant Mccurdy and Stephen Hocking) who put their photos under the Unsplash license~(\url{https://unsplash.com/license}) for public use.

{\small
\bibliographystyle{ieee_fullname}
\bibliography{SMGAN-Arxiv}
}

\end{document}